\documentclass[runningheads]{llncs}
\usepackage{lineno}
\usepackage[T1]{fontenc}
\usepackage{graphicx}
\usepackage{algorithm}
\usepackage{algpseudocode}
\usepackage{amsmath}
\usepackage{amssymb}
\usepackage[misc]{ifsym}
\usepackage[export]{adjustbox}
\usepackage{orcidlink} 

\title{Spatially Grounded Explanations in Vision–Language Models for Document Visual Question Answering}

\titlerunning{Spatially Grounded Explanations for DocVQA} 
\author{
Maximiliano Hormazábal\inst{1,2} \textsuperscript{(\Letter)} \orcidlink{0009-0003-3687-1924} \and
Héctor Cerezo-Costas\inst{2} \orcidlink{0000-0003-2813-2462} \and
Dimosthenis Karatzas\inst{1} \orcidlink{0000-0001-8762-4454}
}

\authorrunning{M. Hormazábal et al.}

\institute{
Computer Vision Center, Universitat Autònoma de Barcelona, Spain \\
\email{\{mhormazabal,dimos\}@cvc.uab.es} \and
Gradiant, Vigo, Galicia, Spain \\
\email{hcerezo@gradiant.org}
}

\begin{document}

\maketitle

\begin{abstract}
We introduce EaGERS, a fully training-free and model-agnostic pipeline that (1) generates natural language rationales via a vision language model, (2) grounds these rationales to spatial sub-regions by computing multimodal embedding similarities over a configurable grid with majority voting, and (3) restricts the generation of responses only from the relevant regions selected in the masked image. Experiments on the DocVQA dataset demonstrate that our best configuration not only outperforms the base model on exact match accuracy and Average Normalized Levenshtein Similarity metrics but also enhances transparency and reproducibility in DocVQA without additional model fine-tuning. Code available at: \url{https://github.com/maxhormazabal/EaGERS-DVQA} 
\end{abstract}

\section{Introduction}

Document Visual Question Answering (DocVQA) has advanced rapidly with Transformer-based methods that integrate OCR, layout modeling, and domain adaptation \cite{appalaraju2021docformer,kim2022ocr,hu2024mplugdocowl2highresolutioncompressingocrfree,powalski2021going}. Concurrently, general-purpose Vision Language Models (VLMs) \cite{9880206,NEURIPS2023_6dcf277e,pmlr-v202-li23q} achieve strong document understanding without explicit DocVQA training.

Deploying off-the-shelf vision–language models in enterprise pipelines often involves costly fine-tuning, unstable prompt engineering and a lack of clear grounding between answers and source regions \cite{zhou2022large}. To address these challenges, we introduce Explanation-Guided Region Selection (EaGERS), a fully model-agnostic, training-free DocVQA pipeline that (i) generates natural language explanations, (ii) selects the top sub-regions over a configurable grid via multimodal embedding similarities and majority voting, and (iii) re-queries the model on a masked image so that answers derive solely from those validated regions ensuring transparency and reproducibility without additional model training.

The problem we address is: how to enforce that the answer can be reconstructed solely from document regions that are explicitly grounded and verbalised, without any additional training of the VLM.

The main contributions of this work are:
\begin{enumerate}
  \item EaGERS: A fully model-agnostic and training-free DocVQA pipeline capable of generating answers on masked document images using general-purpose multimodal models.
  \item Integration of text explanations and visual masking to contribute to the traceability and explainability of inferences.
\end{enumerate}

\section{Related Work}

\subsection{DocVQA}

In recent years, DocVQA systems have achieved solid baselines on the DocVQA dataset \cite{mathew2021docvqa}, some approaches combine OCR and QA modules such LayoutLM \cite{Xu_2020} and TILT, yet still lag behind human performance. Other transformer-based models, such as DocFormer and Donut, adopt end-to-end, OCR-free architectures, and supervised-attention methods like M4C \cite{9156750} integrate textual, positional, and visual cues to boost retrieval accuracy; however, their interpretability remains stuck to attention-weight analysis. Today, state-of-the-art OCR-free approaches reach near-human accuracy but offer no natural language rationales. Meanwhile, multimodal compression frameworks like mPLUG-DocOwl 2.0 improve scalability but still lack explanations grounded in specific document regions.

\subsection{Explainability en DocVQA}

In the area of spatial explainability, methods such as DocXplain \cite{saifullah2024docxplain} apply ablation techniques on document segments to measure their impact on predictions, demonstrating higher fidelity than Grad-CAM \cite{8237336} at the cost of multiple inferences. Hybrid models like DLaVA \cite{mohammadshirazi2024dlavadocumentlanguagevision} combine textual answers with bounding boxes, achieving  Intersection over Union (IoU) \cite{8953982} above $0.5$ on DocVQA, while MRVQA \cite{li2025convincingrationalesvisualquestion} introduces textual rationales alongside visual highlights and proposes specific visual-text coherence metrics.

\subsection{Modal Alignment and Multimodal Embeddings}

Modal alignment projects text and image representations into a shared latent space, allowing direct comparison via geometric metrics such as cosine similarity. Pretrained models like CLIP \cite{pmlr-v139-radford21a}, ALIGN \cite{pmlr-v139-jia21b}, and BLIP \cite{pmlr-v162-li22n} employ a multimodal contrastive objective to bring semantically related pairs closer together. In our pipeline, we use embeddings from BLIP, CLIP, and ALIGN to vectorize both natural language explanations and document sub-regions, enabling the multimodal similarity measurements that drive masking and focused re-querying.

\section{Methodology}

\begin{figure}[h!]
\includegraphics[width=\textwidth,alt={EaGERS Document VQA pipeline that generates a spatial natural language explanation from the image and the question}]{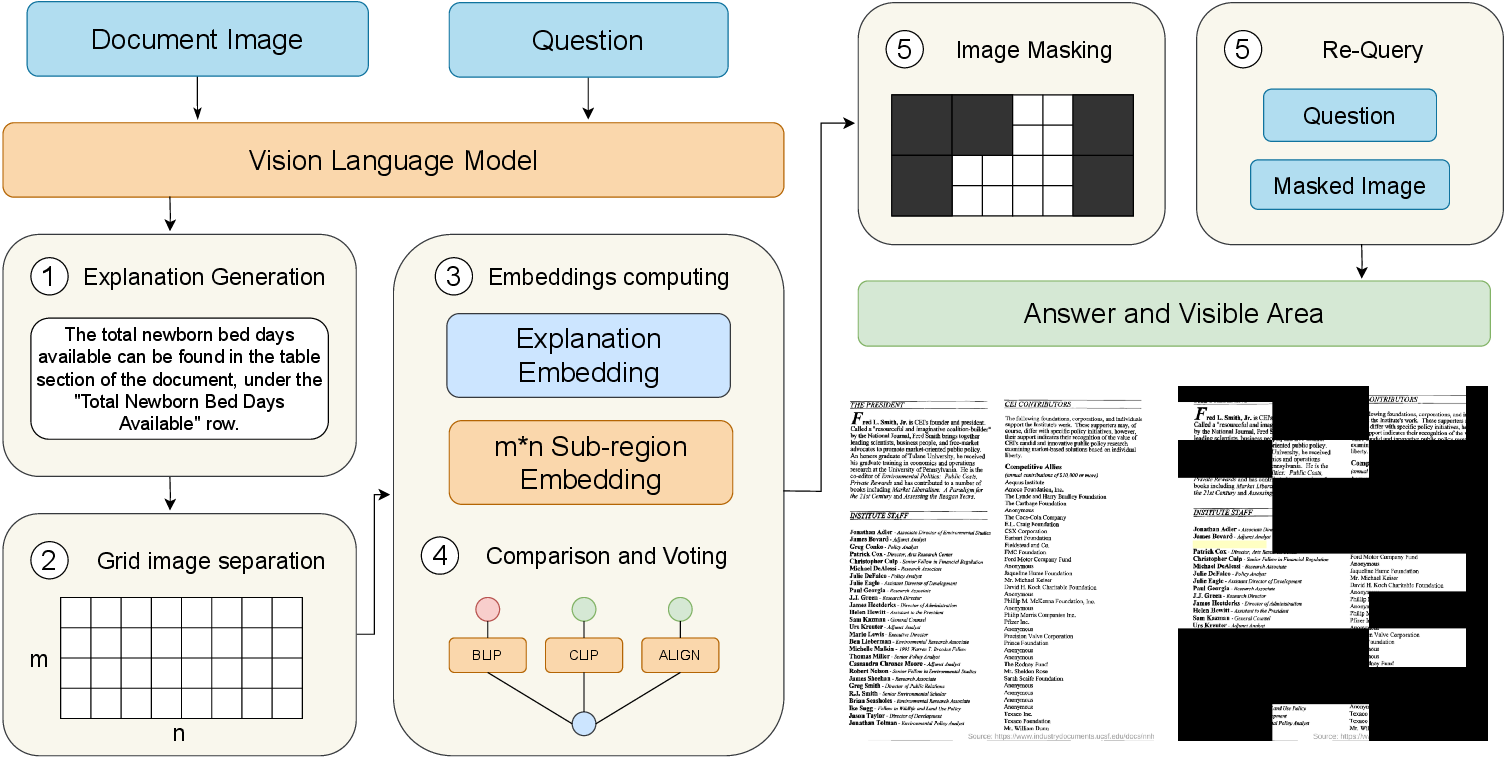}

\caption{EaGERS Document VQA pipeline: (1) the multimodal model generates a spatial natural language explanation from the image and the question; (2) the image is segmented into an $m\times n$ grid; (3) embeddings of the explanation and each sub-region are obtained using BLIP, CLIP, and ALIGN; (4) majority voting selects the most relevant regions; (5) the image is masked to retain only those regions, and the model is re-queried with the question to generate the final answer.}
\label{fig1}
\end{figure}

EaGERS-DocVQA leverages the knowledge of general-purpose multimodal models in document understanding without requiring dedicated training. Figure \ref{fig1} shows the overall architecture of the proposed system, which consists of three main stages: A) Explanation generation, B) Region selection, and C) Answer generation. In our experiments, we use the Qwen2.5VL-3B model as the core component; however, the proposed system is essentially model-agnostic.

\subsection{Explanation Generation}

The document image and question are passed to a vision–language model, which uses them to generate a natural language explanation of how to obtain the requested answer in visual terms. This explanation is not the final answer but serves as a guide to locate the relevant information in the image. We employ this inference in subsequent steps as a semantic tool for comparing the image sub-regions with the generated explanation. Although these spatial explanations typically exhibit good alignment with relevant regions, there are occasional cases where the generated explanations may inaccurately refer to irrelevant areas, potentially impacting subsequent region selection.

\subsection{Region selection via similarity}

The document image is divided into an $m \times n$ grid, yielding $m \cdot n$ sub-regions, each of which may or may not be relevant for obtaining the answer.

Each sub-region is converted into a vector representation in order to compare it with the model’s spatial explanation by an ensemble of three multimodal models: BLIP, CLIP, and ALIGN to generate embeddings in a complementary way in order to mitigate specific biases and improve robustness across heterogeneous document layouts.

Once we obtain a similarity score based on cosine similarity between reasoning and sub-region embeddings for each embedder, we select the top \(k\) for the final answer, where \(k\) is 30\% (rounded up) of sub-regions based on preliminary experiments, as this provided a good compromise between spatial granularity for evidence localization. Although an adaptive grid might better handle irregular layouts, the fixed grid simplifies reproducibility and reduces complexity for this initial study.

The final ranking is determined by majority voting across the rankings from each embedder. In the event of ties during majority voting, we resolve these by prioritizing regions based on their average cosine similarity scores across all embedders, thus favoring sub-regions with more consistent overall relevance. The resulting list of selected sub-regions \(\mathcal{R}\) defines the visible area used in the subsequent answer generation stage.

\subsection{Masking and re-query}

We create a masked version of the original image in which all grid cells outside \(\mathcal{R}\) are filled with black. The question and the masked image are then reintroduced to the same multimodal model, which must generate the answer using only the information within the justified regions, \textbf{without access} to the previously generated explanation except implicitly through the defined region mask.

In the following section we evaluate how effectively this approach recovers ground-truth answers under different grid and margin configurations.

\section{Experimentation}

\subsection{Datasets and Evaluation Protocols}

In our experiments, we use the validation split of the DocVQA Single Page dataset. We applied resizing preprocessing (preserving the aspect ratio) to compress images and optimize inferences. For spatial partitioning, we divide each image into a uniform grid of 5 columns and 5 rows (25 cells) in an initial series of tests, and an alternative configuration of 5 columns and 10 rows (50 cells) to explore the impact of granularity on relevant-region selection. We also have tested a margin expansion of the 15\% of the unmasked sub-region for both amount of cells.

We use Exact Match (EM) and Average Normalized Levenshtein Similarity (ANLS) \cite{60df40bf577646d8aa779dc12132bf5d} which mitigates the impact of misrecognition errors by thresholding normalized edit distances.

\subsection{Main Results}

The table \ref{tab:main_results} presents the performance results (ANLS and EM) of the different pipeline configurations. For comparison we have also run the Qwen2.5-VL-3B model directly in the DocVQA dataset to compare its performance with the pipeline. In addition to this, the unit inference time (model-only and EaGERS pipeline) have been measured to calculate the average inference time for each approach and their coefficients of variation (CV) have been calculated to gain insights of the similarity of times inference between document\footnote{The GPU-hardware used for the experiments were 48GB NVIDIA RTX 6000 Ada Generation.}.

\begin{table}[!h]
  \centering
  \caption{Overall performance (EM and ANLS) on validation for different grid and margin settings.}
  \label{tab:main_results}
  \begin{tabular}{|cccc|cc|cc|}
    \hline
    \multicolumn{4}{|c|}{\textbf{Configuration}} & \multicolumn{4}{c|}{\textbf{Performance}} \\ \hline
    \textbf{Model} & \textbf{Cols} & \textbf{Rows} & \textbf{Margin} & \textbf{EM (\%)} & \textbf{ANLS} & \textbf{Avg Time (s)} & \textbf{CV (\%)} \\ \hline
    $EaGERS_{25|0}$  & 5 & 5  & $0\%$  & 64.20 & 75.25 & 16.98 & 37.08 \\ \hline
    $EaGERS_{50|0}$ & 5 & 10 & $0\%$  & 66.67 & 77.94 & 17.49 & 20.78 \\ \hline
    $EaGERS_{25|15}$ & 5 & 5  & $15\%$ & 72.72 & 82.52 & 16.72 & 24.04 \\ \hline
    $EaGERS_{50|15}$ & 5 & 10 & $15\%$ & \textbf{74.50} & \textbf{83.31} & 17.48 & 20.72 \\ \hline
    Qwen2.5-VL-3B & $\ast$ & $\ast$ & $\ast$ & 71.17 & 82.90 & \textbf{7.21} & 30.08 \\ \hline
  \end{tabular}
\end{table}

As shown, adding a 15\% masking margin consistently improves both EM and ANLS across grid sizes. In particular, the $50$ cells grid with a 15\% margin yields the best performance, suggesting that finer spatial granularity combined with slight overlap enhances the localization and understanding of regions relevant to the answer. This suggests that using a fixed distribution grid across the entire image can introduce complications when measuring cosine distance between embeddings. Specifically, relevant zones may fall on the borders between sub-regions, which could explain discrepancies (such as a margin of $0$ versus $15$) when the grid slightly shifts and either includes or excludes the answer location. This indicates clear future steps in improving the system towards a more flexible grid.

It is also relevant to see that the performance of the model "as-is" has gained explainability without experiencing loss in its performance level, but in the best version $EaGERS_{50|15}$ presents a modest improvement pointing out that restricting the viewing space of the model so that it has more clarity of the answer is an interesting research direction. 

Although this first study does not aspire to a head-to-head comparison with the state-of-the-art methods yet. It is also relevant to comment that EaGERS is above the solutions initially proposed in DocVQA \cite{mathew2021docvqa} which have shown ANLS results in VQA models such as LoRRA=$0.110$ or M4C=$0.385$, BERT QA systems around $0.655$ and multimodal architectures such as LayoutLMv2-BASE with an ANLS of $0.7421$. However, next steps will be to evaluate this model against state-of-the-art solutions, which achieve even higher ANLS values.

\section{Limitations and Future Work}

Among the limitations of this study that suggest promising directions for future work: relying on fixed grid configurations may not generalize well to documents with irregular layouts or variable aspect ratios; object detectors may be useful for subdividing relevant regions \cite{GOMEZ2021242}. Another important limitation is the dependence on spatial explanations generated by VLMs, which may occasionally produce inaccurate rationales, leading to incorrect region selections. Our pipeline assumes spatial accuracy of VLM explanations; however, when these do not match ground truth, the final fidelity may degrade which an aspect we will address in future work.

Future work should include systematic evaluations of the frequency and impact of such inaccuracies on the final results, we will evaluate more robust fusion strategies, such as Reciprocal Rank Fusion (RRF), and quantify agreement using Krippendorff alpha ($\alpha$), in order to analyze in depth the internal consistency of spatial selections. 

It would also be valuable to extend experiments to datasets that explicitly involve answer-localization annotations, enabling the use of quantitative fidelity metrics, such as (IoU) or visual–text coherence scores. Moreover, efficiency improvements are needed considering the increase in inference time reported in Table \ref{tab:main_results}. Finally, we plan to explore methods to assess properly the level of explainability in comparison with alternatives.

\section{Conclusions}

We have proposed a pipeline that unifies natural language explanations with quantitative region selection and masked re-querying to ensure answers derive only from validated document regions. Our approach yields significant improvements in Exact Match and ANLS over standard baselines, demonstrating enhanced transparency and reproducibility. In future work, we will investigate adaptive grid partitioning to better handle structural variability, conduct comprehensive ablation studies to optimize embedder selection, and directly compare performance with more advanced state-of-the-art models. We also plan to integrate quantitative explainability metrics and carry out user studies to assess the clarity and reliability of the generated explanations.

%
%
%
\bibliographystyle{splncs04}
\bibliography{mybibliography}

\end{document}